\documentclass[journal]{IEEEtran}
\ifCLASSINFOpdf
\else
\fi
\usepackage{amsmath}
\usepackage{graphicx}
\usepackage[colorlinks,linkcolor=red]{hyperref}
\usepackage{amssymb}
\usepackage{threeparttable}
\usepackage{tikz}
\usepackage{amsthm}
\usepackage{mathrsfs}
\begin{document}
%
\title{MSFNet:Multi-scale features network for monocular depth estimation}

\author{Meiqi Pei
\thanks{

\textit{(Corresponding author: Yunzhou Zhang.)}
}
}

\markboth{Journal of \LaTeX\ Class Files,~Vol.~14, No.~8, August~2015}%
{Shell \MakeLowercase{\textit{et al.}}: Bare Demo of IEEEtran.cls for IEEE Journals}


%


\maketitle

\begin{abstract}
In recent years, monocular depth estimation is applied to understand the surrounding 3D environment and has made great progress. However, there is an ill-posed problem on how to gain depth information directly from a single image. With the rapid development of deep learning, this problem is possible to be solved. Although more and more approaches are proposed one after another, most of existing methods inevitably lost details due to continuous downsampling when mapping from RGB space to depth space. To the end, we design a Multi-scale Features Network (MSFNet), which consists of Enhanced Diverse Attention (EDA) module and Upsample-Stage Fusion (USF) module. The EDA module employs the spatial attention method to learn significant spatial information, while USF module complements low-level detail information with high-level semantic information from the perspective of multi-scale feature fusion to improve the predicted effect. In addition, since the simple samples are always trained to a better effect first, the hard samples are difficult to converge. Therefore, we design a batch-loss to assign large loss factors to the harder samples in a batch. Experiments on NYU-Depth V2 dataset and KITTI dataset demonstrate that our proposed approach is more competitive with the state-of-the-art methods in both qualitative and quantitative evaluation.
\end{abstract}



%
\IEEEpeerreviewmaketitle

\section{Introduction}
\IEEEPARstart{M}{onocular} depth estimation, whose sensor has the advantages of low cost and easy calibration, becomes a more economical and flexible solution for depth acquisition and has an unshakable position in fields such as autonomous driving and robotics. However, understanding 3D perception information from an image is an ill-posed problem~\cite{zhao2020monocular}. The reason is that the result of projecting from 3D space to 2D space is certain, on the contrary,  projecting from  2D  space to  3D  space has multiple results. With the rapid development of convolutional neural networks (CNN), monocular depth estimation based on deep learning has become a research hotspot and gained significant progress.

Since Eigen \textit{et al}.~\cite{2014Depth} firstly used convolutional neural networks for depth estimation tasks, various deformed network structures~\cite{shelhamer2015scene, mancini2016fast, wofk2019fastdepth, zhang2015monocular, li2017two, narihira2015learning} have gradually been applied in this vision task. Due to operations such as convolution and pooling in CNN reducing the resolution of the feature map, detailed information will inevitably be lost. Many researchers have discovered this phenomenon and arranged research schemes. Initially, Eigen \textit{et al}.~\cite{2014Depth} utilized the coarse branch and the fine branch to complement each other with global information and local detailed information. Unlike the~\cite{2014Depth}, Laina \textit{et al}.~\cite{laina2016deeper} designed a fast upsampling module to continuously learn detailed information under the supervision of loss. Compared with the downsampling-upsampling one-line structure like~\cite{laina2016deeper} to restore information, Fu \textit{et al}.~\cite{fu2018deep} thought of employing atrous spatial pyramid pooling (ASPP) to fuse information among features at the same stage to obtain better-predicted results. Afterward, Hu \textit{et al}.~\cite{hu2019revisiting} also considered feature fusion, but tried bilinear interpolation for upsampling in features at different stages and then fused with the output of the decoder. In addition to the loss of downsampling information leading to deviations in the predicted depth maps, the ingenious design of the loss function also has an impact on final result. For example, Padhy \textit{et al}.~\cite{padhy2019multi} make full use of the features among different stages to complement each other, since lacking more accurate loss functions, the effect was seriously reduced.

\begin{figure}    
\centering    
\includegraphics[width=87mm]{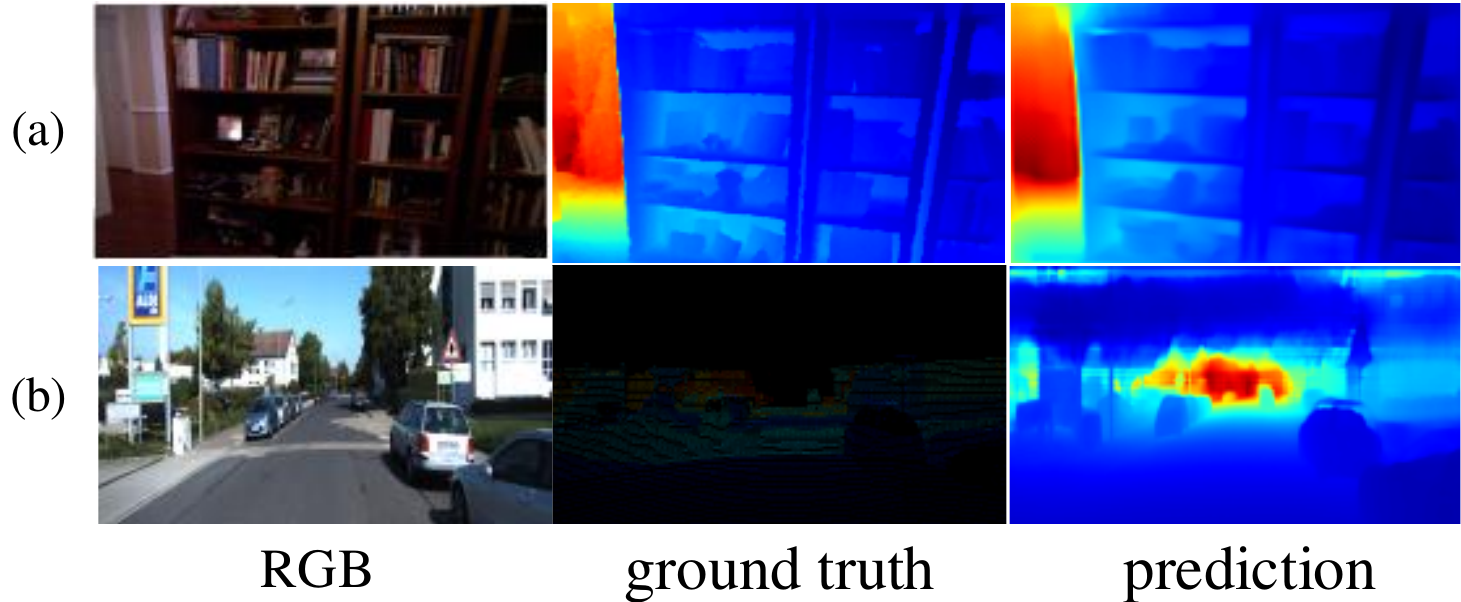}\\    
\caption{Examples of estimated depth maps. The top row is the result of NYU-Depth V2 dataset, while the bottom row is the result of KITTI dataset. The three columns from left to right are RGB input, ground truth, and the result of our method.}
\label{Fig1}      
\end{figure} 

In this work, considering the issue mentioned above that the convolution and pooling operations in the downsampling process will lose information, which cannot be recovered by simple upsampling blocks, we propose a Multi-scale Features Network (MSFNet). The main idea is to add detailed information from the underlying stage during the upsampling process to optimize the predicted depth map. In MSFNet, there are two important components, named Enhanced Diverse Attention (EDA) module and Upsample-stage fusion (USF) module. By means of spatial attention, the EDA module obtains feature maps with efficient spatial information in the high-level stage. Considering the influence of noise in the network training process and subtracting the high responses making the network dig out more significant information in the remaining areas, the EDA module achieves more robust spatial attention information through utilizing the minimum pooling. In response to the problem of losing detailed information, we put forward the USF module to merge the features extracted from multiple stages, which contains the low-level detailed information and the high-level semantic information. At the same time, the USF module also refers to efficient sub-pixel convolution proposed by Shi \textit{et al}.~\cite{shi2016real}. Compared with the bilinear interpolation, the efficient sub-pixel convolution not only improves the speed of upsampling process but also restores high-resolution feature maps. In the supervised monocular depth estimation task, the loss function is generally applied to supervise a single image during the training process. We design a loss function that considers the difference among samples, named batch-loss, to enable the network to converge more quickly. Specifically, the samples with larger loss values are given larger weights, while the samples with smaller loss values are worth smaller weights. In the ablation studies section, it is testified the effectiveness of the loss function. The prediction results on NYU-Depth V2 dataset and KITTI dataset using the complete MSFNet structure are shown in Fig.\ref{Fig1}.

To sum up, Our contributions are as follows:
\begin{itemize}
    \item We propose a Multi-scale Features Network, which can more forcefully enrich the detailed information of the feature map during the upsampling process, to obtain better prediction results.
    \item We design an Enhanced Diverse Attention module to calculate feature maps with spatial attention, so that not only neighboring pixels, but also the responses of other associated regions are used when restoring the feature maps.
    \item Our Upsample-Stage Fusion module collects feature information from multiple stages, fully integrating these feature maps through a novel structure instead of simple cascading, and utilizes the efficient sub-pixel convolution to receive high-resolution prediction.
    \item Our batch-loss, which considers the loss of different samples, calculates the distance between the prediction and the ground truth more effectively from the perspective of the sample.
    \item The proposed method has been verified on the challenging NYU-Depth V2 dataset and KITTI dataset and achieved excellent performance compared with many recent methods.
\end{itemize}

\section{Related Work}
 Gaining depth information from an image is a basic but critical visual task. Perceiving the 3D structure of the scene based on the depth information is a key part of the task performed by the artificial intelligence system. By recognizing the surrounding environment, the control center can continuously exchange information with the outside world, to realize functions such as visual ranging and target positioning in the decision-making process. Initially, multi-view image pairs were used to get depth information based on geometric stereo matching. SFM (Structure-from-Motion)~\cite{7785097} method is commonly employed in a single view. Nonetheless, the above methods require extra information in addition to a single image such as relative transformations in multi-views and motion information in single-views. And it was still difficult to exactly recover depth information with only one RGB image.

 With the rapid development of deep learning, the application of CNN for monocular depth estimation has become convenient and potent. Eigen \textit{et al}.~\cite{2014Depth} first utilized CNN for monocular depth estimation tasks, making it possible to obtain depth information from a single image. However, they applied the fully connected layer in the coarse branch and then upsample, which inevitably caused serious errors in the recovery process of the global information. With the excellent performance of FCN~\cite{long2015fully} in semantic segmentation tasks, full convolution is gradually adopted instead of a fully connected layer. And there is no limit to the input image size of the network so that the feature map can exhibit adequate global information without becoming a particularly small size, which is a breakthrough for pixel-level vision tasks.

 After applying the encoder-decoder structure in computer vision tasks, more and more researchers have discovered the drawbacks of the encoder structure. Although the encoder can effectively extract a large range of detailed information and the hidden semantic information in different stages, some necessary information is inevitably lost in the process of convolution and pooling. Therefore, many works have studied how to add supplements to the underlying information in the decoder. U-Net~\cite{ronneberger2015u} adopted the skip connection in the network structure to straightly connect the features derived from the encoder to the decoder and the restored features were more cogent than the step-by-step restoration directly from the decoder. The work of Padhy \textit{et al}.~\cite{padhy2019multi} was a typical method of depth estimation with the skip connection. According to it, the baseline structure we adopt also applies the skip connection, but the prediction accuracy of the network structure that only utilizes the skip connections needs to be improved.

 Most object detection tasks employed multi-scale structures to predict small objects~\cite{lin2017feature}, while many studies restored detailed information by multi-scale structures in monocular depth estimation tasks. Li \textit{et al}.~\cite{li2017single} selected different levels of features, by cascading and dropout to merge features. Xu \textit{et al}.~\cite{xu2017multi} adopted continuous CRF to integrate multi-scale output feature maps. Fu \textit{et al}.~\cite{fu2018deep} utilized ASPP to obtain feature maps of different scales at the same stage and then concatenated them. Hu \textit{et al}.~\cite{hu2019revisiting} directly sampled the features at different stages to the same size by bilinear interpolation. Chen \textit{et al}.~\cite{DBLP:conf/ijcai/ChenCZ19} exploited the pyramid fusion method for the characteristics of the pyramid structure, and then gradually upsampled features to restore to the original image size. The structure was valid, but the process was too complicated. The fusion method we propose can simply and effectively avail of the information of each stage and merge it. 
 \begin{figure*}    
\centering    
\includegraphics[width=170mm]{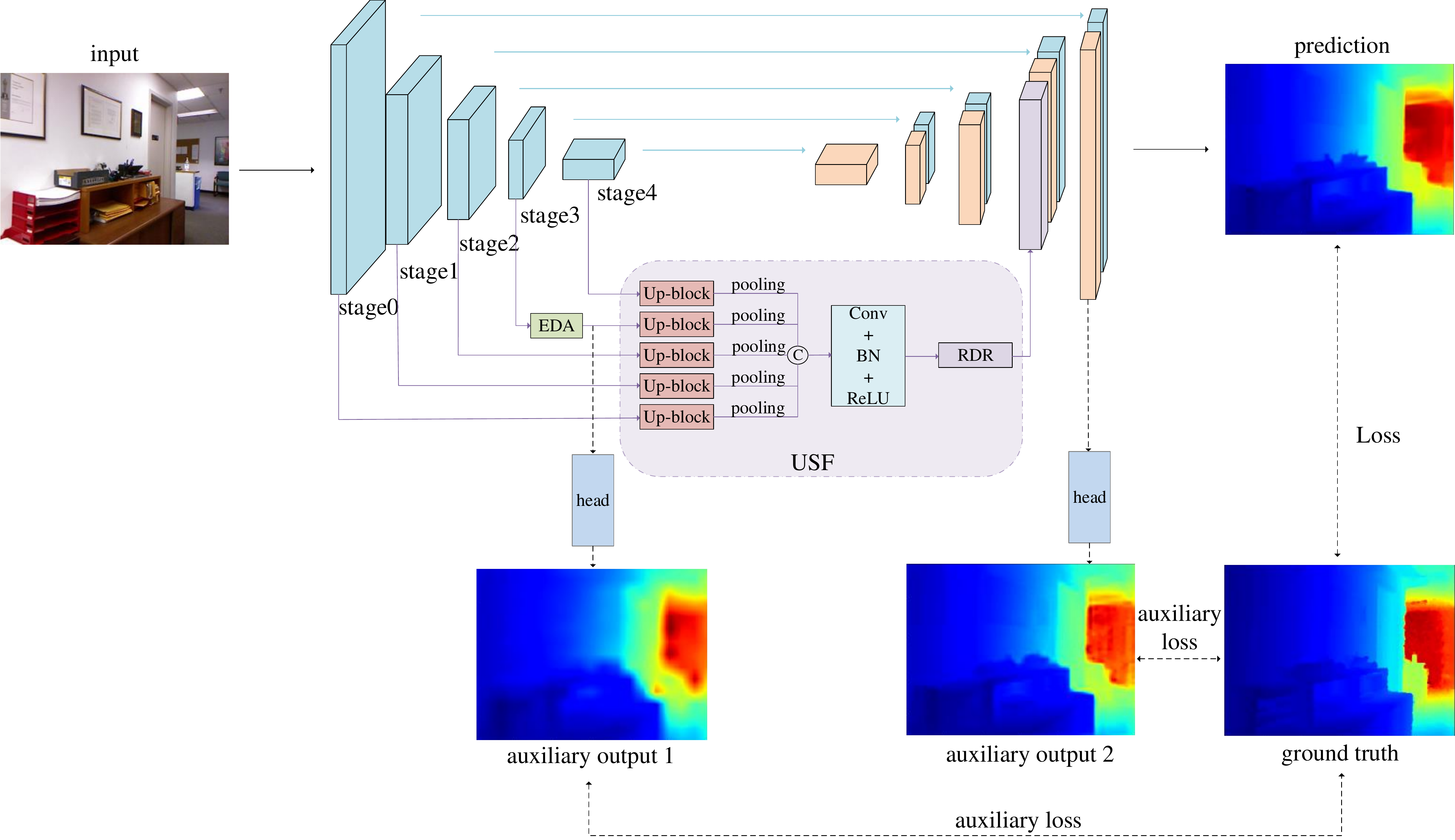}\\    
\caption{An overview of the Multi-scale Features Network (MSFNet). The overall network structure is the encoder-decoder structure. The blue blocks in the figure are the features of different stages in the encoding process, and the orange blocks are the features gotten by upsampling the output of the previous stage. In the decoding process, we add our EDA module (shown in the green block) and USF module (shown in the purple dashed box). Two heads are exploited at the output of the EDA module and the next stage of the USF module to reduce the high-dimensional features to one dimension, which is convenient for calculating the auxiliary loss and converging the model faster during the training process.}
\label{Fig2}    
\end{figure*} 
Meanwhile, these methods fused features based on the information extracted by the encoder which decides the richness of the information. Liu \textit{et al}.~\cite{liu2019fully} and Su \textit{et al}.~\cite{su2020monocular} chose the channel attention mechanism for enriching the semantic information of the features, but for spatial details, it was not improved well. In contrast, We reckon that utilizing the spatial attention mechanism can obtain outstanding performance by mining the spatial details.

\section{Methodology}
The MSFNet consists of three parts: an encoder structure, an enhanced diverse attention module, and an upsample-stage fusion module. We primarily introduce the complete architecture of the network in Sec.\ref{A}. Then the EDA module and the USF module are recommended in detail respectively in Sec.\ref{B} and Sec. \ref{C}. The batch-loss and other loss functions utilized in this paper are suggested in Sec.\ref{D} at last.
\subsection{Multi-scale Features Network}
\label{A}

The overall network architecture is shown in Fig. \ref{Fig2}. In the encoding process, we utilize the SENet154~\cite{hu2018squeeze}, an extensive adoption in most previous works~\cite{DBLP:conf/ijcai/ChenCZ19, hu2019revisiting, durasov2019double}, as the backbone of our encoder with a minor modification. We only retain the feature extraction part and discard the following fully connected layer. Given an input image with size $W \times H$, the size of feature maps are gradually reduced by half through each stage. Therefore, the feature map is captured by the highest layer with size $W/32 \times H/32$. Considering that there is detailed information on the low-level stage and semantic information on the high-level stage, we make most of the low-level and high-level multi-scale features to mitigate the loss of detailed information in the next.


The decoder consists of basic blocks of convolutional layers. And they are applied after the concatenation of the previous block through the 2$\times$ bilinear interpolation with the block in the encoder with the same spatial size. It has proven to be an effective strategy in~\cite{alhashim2018high}. Note that we utilize the Leaky ReLU to activate features and $\alpha$ is also set to 0.2 followed~\cite{alhashim2018high} in our experiment.


\begin{equation}
f(x)=
\left\{
\begin{array}{rcl}
x, & &\text{if  x \textgreater 0}\\ 
\alpha x, & & \text{otherwise}
\end{array} \right.
\end{equation}

Meanwhile, we propose an EDA module and a USF module. The output of the stage3 is sent into the EDA module to generate more diverse detailed information for the USF module. For the USF module, we employ the outputs of the EDA module and another four stages of the encoder backbone as input and merge more details and semantic information for the decoder. More details can be found in Section~\ref{B} and Section~\ref{C}, respectively.



\subsection{Enhanced Diverse Attention Module}
\label{B}
\begin{figure}    
\centering    
\includegraphics[width=\linewidth]{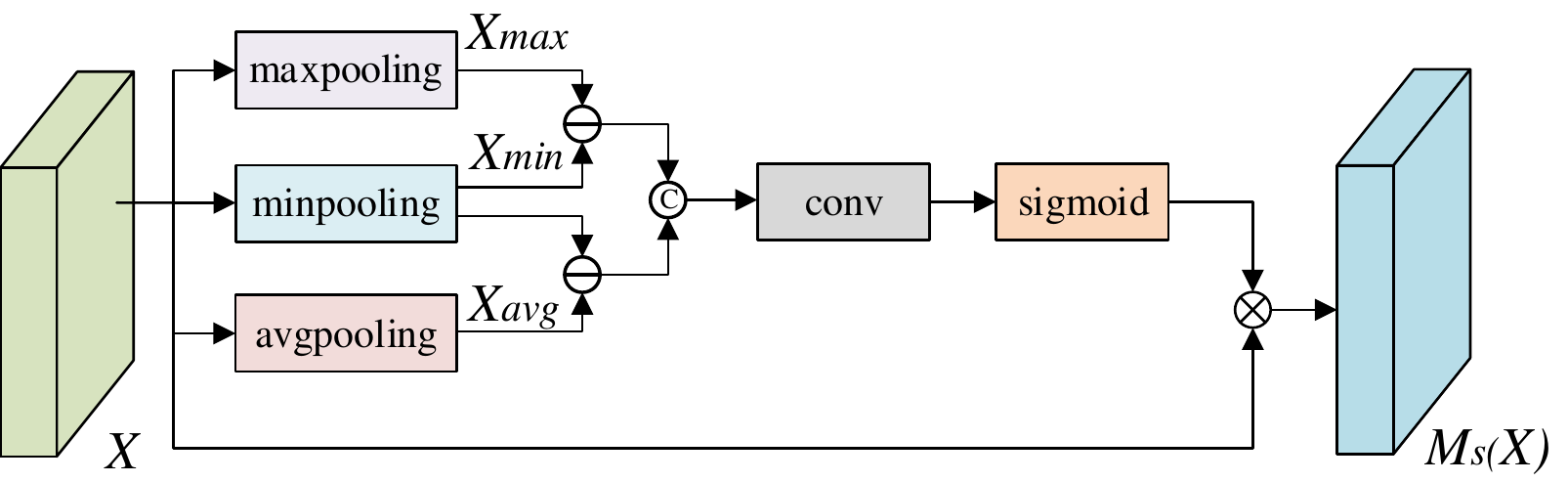}\\    
\caption{The framework of enhanced diverse attention module. $\ominus$ denotes the element-wise subtraction, $\textcircled{c}$ represents the concatenation operation along with the channel and $\otimes$ is the element-wise product.}
\label{Fig3}    
\end{figure}

Recently, attention mechanisms have been attractive as aim at strengthening crucial features and suppressing irrelevant ones, which is an effective method in excavating more detailed information. Based on the above introduction, we also design an effective attention module to retain as much detailed information as possible from encoder to decoder. Inspired by the work~\cite{woo2018cbam} which proposed the spatial attention module of convolutional block attention module (CBAM-S), we propose an enhanced diverse attention (EDA) module in this paper. The architecture of the EDA module is shown in Fig. \ref{Fig3}. 

Specifically, given the input feature map $X$, we apply the maximum pooling (maxpooling), minimum pooling (minpooling), and average pooling (avgpooling) along the channel dimension to generate the $X_{max}$, $X_{min}$, $X_{avg}$. The EDA module is formulated as follows,
\begin{equation}
M_s(X) = \sigma(f^{7 \times 7}([X_{avg}-X_{min}; X_{max}-X_{min}])) * X
\end{equation}

where $\sigma(\cdot)$ is the sigmoid operation, and $f^{7 \times 7}$ is the convolution operation with the convolution kernel of 7. The significant difference is that we apply minpooling to obtain minimum feature value and subtract the minimum value from the maximum and average values separately. The main reasons for adding minpooling are as follows: 1) The noise along the channel dimension should be linear for the maximum and minimum area after convolution, and the minimum is subtracted to tellingly suppress noise interference. 2) Since the effective regions of spatial attention focus on will be weakened after subtracting the minimum feature value, the model will have to pay attention to more effective regions. Therefore, the EDA module can assist the model to learn more detailed information, and thus guarantee better performance. 

It is worth noting that we only embed the EDA module between stage3 and stage4 to extract the feature, and then send it into the USF module. This is because the output of stage3 simultaneously contains more details and semantic information among other stages.

\subsection{Upsample-Stage Fusion Module}
\label{C}
In addition to using the EDA module to obtain more detailed information, an effective feature fusion strategy can also alleviate the loss of detailed information. Hu \textit{et al}.~\cite{hu2019revisiting} proposed the MFF structure, in which features from different stages were firstly unified to the same scale with bilinear interpolation and then cascaded. Unlike the simple cascade operation of~\cite{hu2019revisiting}, we design a more novel and effective feature fusion method named upsample-stage fusion (USF) module as shown in the purple region in Fig.~\ref{Fig2}. Specifically, the output of the EDA module and the other four stages are sent into different Up-block proposed in~\cite{shi2016real} to increase resolution. And then the adaptive pooling ( "pooling" in Fig.~\ref{Fig2} ) operations are used to resize the feature to the same size. After that, the convolutional block (the blue rectangular in USF module) followed by the adaptive pooling is applied to integrate the feature further. Finally, we send the feature into the Reduced Dimension Refinement (RDR) module to generate more effective information for the decoder.

\begin{figure}    
\centering    
\includegraphics[width=87mm]{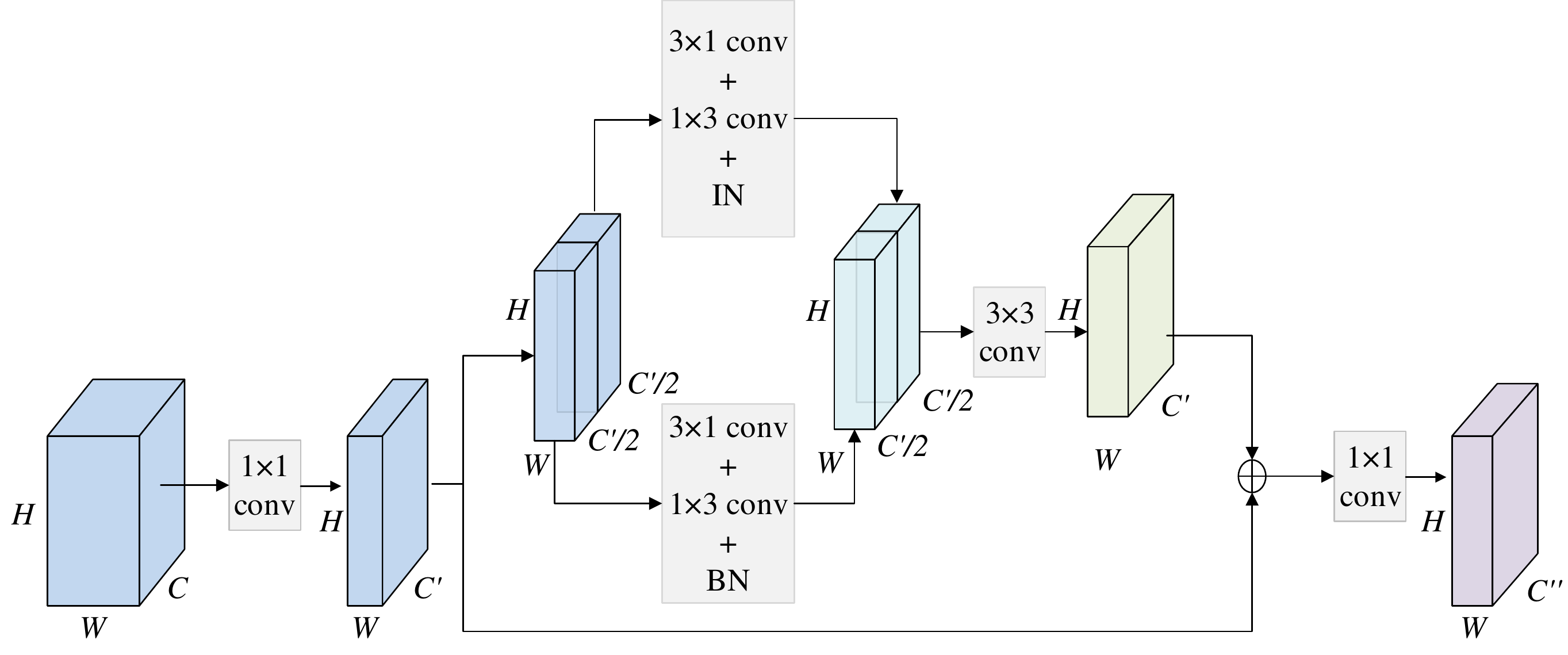}\\    
\caption{Diagrams of the Reduced Dimension Refinement module.}
\label{Fig5}    
\end{figure}

More details about the RDR module are shown in Fig.\ref{Fig5}. In the RDR module, the input tensor is first squeezed with a $1\times1$ convolution to extract valid information among channels. Then, because instance normalization (IN) and batch normalization (BN) can learn the invariant appearance and content-related information, respectively, we utilize the IN and BN module to reinforce features, which plays a critical role in monocular depth estimation prediction. We describe the detailed operation below. The tensor from the output of the $1\times1$ convolution is divided equally into two parts along the channel axis. After that, these two parts of features are sent into IN and BN module to generate features including invariant appearance and content-related information, respectively. Note that we apply the separable convolution to turn the original $3 \times 3$ convolution into a $3 \times 1$ convolution and a $1 \times 3$ convolution, which can not only improve the feature but also reduce the number of parameters. The following tensors from the IN and BN modules are concatenated and convolved by a $3\times3$ convolution layer, producing the improved tensor. Finally, we integrate the improved tensor with the original input tensor and apply a $1\times1$ convolution layer to generate the refinement feature, i.e, purple rectangle in Fig.~\ref{Fig5}.  

\subsection{Loss Function}
\label{D}
To predict the depth information in the scene accurately, we utilize multiple loss functions to train the network correctly. The overall loss function, consisting of four items, is defined by
\begin{equation}
\begin{aligned}
& L(y,\hat{y})=\\
& \lambda L_{batch}(y, \hat{y}) + L_{grad}(y, \hat{y})+L_{SSIM}(y, \hat{y})
 +\mu L_{aux}(y_2,y_3,\hat{y})
\end{aligned}
\end{equation}

where $y$ is the predicted depth map, $\hat{y}$ represents the ground truth. $y_2$ and $y_3$ are the feature maps from the EDA module and USF module, respectively. $\lambda$ and $\mu$ are the weighting factors.

The loss $L_{batch}$, batch-loss proposed in this paper, is used to automatically weaken the contribution of easy examples during training and rapidly focus the model on hard examples in a batch, which is beneficial to achieve better performance. Note that the original $L_1$ loss is also added in the batch-loss for converging the model faster. The batch-loss is defined as follow:
\begin{equation}
    L_{batch}(y,\hat{y})=\frac{1}{n}\sum_{i=1}^{n} w_i * (|y_i-\hat{y_i}|)+\frac{1}{n}\sum_{i=1}^{n}|y_i-\hat{y_i}|
\end{equation}
where $n$ is the number of all pixels with the depth value of an image and $w_i=\frac{e^{|y_i-\hat{y_i}|}}{\sum_{i=1}^{n}e^{|y_i-\hat{y_i}|}}$ is the modulating factor obtained by the result of each sample’s the deviation in a batch. We note the property of the batch-loss. As the gap between prediction and ground truth is increased, the effect of the modulating factor is likewise increased, and in turn, the loss function down-weights easy examples.

The second loss $L_{grad}$ calculates the gradient $g$ information of the depth image:
\begin{equation}
    L_{grad}(y,\hat{y})=\frac{1}{n}\sum_{i=1}^{n}|g_x(y_i,\hat{y_i})|+|g_y(y_i,\hat{y_i})|
\end{equation}
where $g_x$, $g_y$, respectively, caculated the differences in the x and y components for the depth image gradients of $y$ and $\hat{y}$.

The third loss $L_{SSIM}$ is adopted to calculate the structural similarity between the predicted result and the ground truth, which is based on the brightness and contrast of the local pattern and similar to the human visual system. The performance of $L_{SSIM}$ has been proved in the work of Alhashim \textit{et al}.~\cite{alhashim2018high} and the same loss function form is utilized as follows:
\begin{equation}
    L_{SSIM}(y,\hat{y})=\frac{1-SSIM(y,\hat{y})}{2}
\end{equation}

The last loss is the auxiliary loss function. We connect a head after the EDA module and USF module, respectively. The head, a $1 \times 1$ convolution, is employed to compress the feature map into a 2D tensor which processed by bilinear upsampling is adopted to calculate auxiliary loss with ground truth. The auxiliary loss consists of $L_1$ loss and can be defined as:
\begin{equation}
    L_{aux}(y_2,y_3,\hat{y})=\frac{1}{n}\sum_{i=1}^{n}|y_{2i}-\hat{y_i}|+\frac{1}{n}\sum_{j=1}^{n}|y_{3j}-\hat{y_j}|
\end{equation}

\section{Experiments}
To illustrate the performance of the proposed network structure and loss function, we conduct experiments on NYU-Depth V2 dataset and KITTI dataset. In addition, qualitative and quantitative comparisons are reported with some state-of-the-arts methods, and the results prove the excellent performance of our method.

We conduct experiments using Pytorch on two NVIDIA Titan XP GPUs with 12 GB memory. SENet-154 pre-trained on the ImageNet dataset is employed as the backbone, and the weights of the rest of the MSFNet are initialized by Xavier~\cite{glorot2010understanding}. In all experiments, the learning rate is initially set to 0.0001, reduced to 5\% every 5 epochs, and the Adam algorithm with a weight decay of 1e-4, $\beta_1=0.9$ and $\beta_2=0.999$, are utilized as the optimizer to update the network.
The batch size is 16 for NYU-Depth V2 dataset while 8 for KITTI dataset. The parameters in the loss function are assigned to $\lambda$=$\mu$=0.1 for NYU-Depth V2 dataset and $\lambda$=$\mu$=1 for KITTI dataset. We train the network for 20 epochs. Data augmentation methods, including random horizontal flip, normalization, and random channel permutations, are adopted to enhance the generalization ability of the model. And Alhashim \textit{et al}.~\cite{alhashim2018high} have confirmed that the color channel augmentation can extremely improve the performance of the network.

We evaluate our network following the several errors which have been used in the prior works~\cite{2014Depth}.

\begin{itemize}
    \item Root mean squared error (RMSE):
    $\sqrt{\frac{1}{|T|}\sum_{y\in T}||y-\hat{y}||^2}$
    \item Mean absolute relative (REL) error: 
       $\frac{1}{|T|}\sum_{y\in T}|y-\hat{y}|/\hat{y}$.
    \item Threshold: percentage of predicted pixels, such that
    $max((\frac{y_i}{\hat{y_i}}),(\frac{\hat{y_i}}{y_i}))=\delta < thr$.
\end{itemize}

$T$ indicates all valid pixels of an image indexed by $i$.

\begin{table*}[]
\caption{Depth estimation results on NYU-Depth V2 Dataset (Boldface denotes the best results, -:not available, $\downarrow$ means the metric lower is better, while $\uparrow$ means the metric higher is better.hereinafter)}
\label{Tab1}
\centering
\begin{threeparttable}
\begin{tabular}{|c|c|c|c|c|c|c|c|}
\hline
                          Method  &  Publication  & REL $\downarrow$   & RMSE $\downarrow$ & log10 $\downarrow$ & $\delta \textless 1.25\uparrow$     & $\delta \textless 1.25^2 \uparrow$     & $\delta \textless 1.25^3 \uparrow$     \\ \hline
Eigen \textit{et al}.~\cite{2014Depth}    & NIPS2014    & 0.215 & 0.907 & -      & 0.611 & 0.887 & 0.971 \\ \hline
Liu \textit{et al}.~\cite{liu2015deep}    & CVPR2015     & 0.23  & 0.824 & 0.095 & 0.614 & 0.883 & 0.971 \\ \hline
Chakrabarti \textit{et al}.~\cite{DBLP:conf/nips/ChakrabartiSS16} & NIPS2016 & 0.149 & 0.62  & -      & 0.806 & 0.958 & 0.987 \\ \hline
Laina \textit{et al}.~\cite{laina2016deeper}   & 3DV 2016     & 0.127 & 0.573 & 0.055 & 0.811 & 0.953 & 0.988 \\ \hline
Cao \textit{et al}.~\cite{cao2017estimating}    & TCSVT2017     & 0.232 & 0.819 & 0.091 & 0.646 & 0.892 & 0.968 \\ \hline
Li \textit{et al}.~\cite{li2017two}   & ICCV2017       & 0.143 & 0.635 & 0.063 & 0.788 & 0.958 & 0.991 \\ \hline
Xu \textit{et al}.~\cite{xu2017multi}    & CVPR2017       & 0.121 & 0.586 & 0.052 & 0.811 & 0.954 & 0.987 \\ \hline
Lee \textit{et al}.~\cite{lee2018single}   & CVPR2018      & 0.139 & 0.572 & -      & 0.815 & 0.963 & 0.991 \\ \hline
Qi \textit{et al}.~\cite{qi2018geonet}   & CVPR2018       & 0.128 & 0.569 & 0.057 & 0.834 & 0.96  & 0.99  \\ \hline
Zhang \textit{et al}.~\cite{zhang2019pattern}    & CVPR2019   & 0.121 & 0.497 & -      & 0.846 & 0.968 & 0.994 \\ \hline
Wang \textit{et al}.~\cite{wang2020sdc}   & CVPR2020     & 0.128 & 0.497 & -      & 0.845 & 0.966 & 0.990  \\ \hline
Su \textit{et al}.~\cite{su2020monocular}   &TITS2020      & 0.137 & 0.498 & 0.058      & 0.826 & 0.967 & \textbf{0.995}  \\ \hline
Wang \textit{et al}.~\cite{wang2020cliffnet}    & ECCV2020    & 0.128 & \textbf{0.493} & -      & 0.844 & 0.964 & 0.991 \\ \hline
ours       &-                 & \textbf{0.118} & 0.531 & \textbf{0.051} & \textbf{0.865} & \textbf{0.975} & 0.993 \\ \hline
\end{tabular}
    \end{threeparttable}
\end{table*}
\vspace{0.2in}
\begin{table*}[]
\caption{Depth estimation results on KITTI Dataset}
\label{Tab2}
\centering
\begin{tabular}{|c|c|c|c|c|c|c|c|}
\hline
Method     & Publication             & REL $\downarrow$  & RMSE $\downarrow$ & $\delta \textless 1.25 \uparrow $     & $\delta \textless 1.25^2 \uparrow $     & $\delta \textless 1.25^3 \uparrow $     \\ \hline
Ladicky \textit{et al}.~\cite{ladicky2014pulling} & CVPR2014 & -      & -      & 0.47  & 0.721 & 0.854 \\ \hline
Eigen \textit{et al}.~\cite{2014Depth} & NIPS2014   & 0.19  & 7.156 & 0.692 & 0.899 & 0.967 \\ \hline
Garg \textit{et al}.~\cite{garg2016unsupervised}  & ECCV2016  & 0.169 & 5.104 & 0.74  & 0.904 & 0.962 \\ \hline
Godard \textit{et al}.~\cite{godard2017unsupervised} & CVPR2017 & 0.148 & 5.927 & 0.803 & 0.922 & 0.964 \\ \hline
Xu \textit{et al}.~\cite{xu2017multi}   & CVPR2017   & 0.125 & 4.685 & 0.816 & 0.951 & 0.983 \\ \hline
Xu \textit{et al}.~\cite{xu2018structured}  & CVPR2018     & 0.122 & 4.677 & 0.818 & 0.954 & 0.985 \\ \hline
Liu \textit{et al}.~\cite{liu2019fully}  & IET2019   & 0.127 & 4.977 & 0.838 & 0.948 & 0.98  \\ \hline
Fang \textit{et al}.~ \cite{fang2020towards}  & WCACV2020  & 0.098 & 4.075 & 0.889 & 0.963 & 0.985 \\ \hline
ours     &-               & \textbf{0.098} & \textbf{4.054} & \textbf{0.893} & \textbf{0.968} & \textbf{0.987} \\ \hline
\end{tabular}
\end{table*}
\vspace{0.2in}
\begin{table*}[]
\caption{Detailed performance comparison of our proposed architecture on NYU-Dpeth V2 subdataset}
\label{Tab3}
\centering
\begin{tabular}{|c|c|c|c|c|c|c|}
\hline
method                      & REL $\downarrow$           & RMSE $\downarrow$          & log10 $\downarrow$            & $\delta \textless 1.25 \uparrow $     & $\delta \textless 1.25^2 \uparrow $     & $\delta \textless 1.25^3 \uparrow $              \\ \hline
baseline                    & 0.21           & 0.717          & 0.081          & 0.693          & 0.918          & 0.977          \\ \hline
baseline+USF                & 0.197          & 0.728          & 0.079          & 0.704          & 0.926          & 0.982          \\ \hline
baseline+USF+CBAM-S           & 0.189          & 0.752          & 0.078          & 0.713          & 0.928          & 0.982          \\ \hline
baseline+USF+EDA            & 0.176          & 0.703          & 0.073          & 0.737          & 0.935          & 0.985          \\ \hline
baseline+USF+EDA+batch-loss & \textbf{0.173} & \textbf{0.658} & \textbf{0.071} & \textbf{0.748} & \textbf{0.945} & \textbf{0.989} \\ \hline
\end{tabular}
\end{table*}

\subsection{Datasets}
NYU-Depth V2 dataset~\cite{silberman2012indoor} is a large indoor scene dataset collected by Microsoft Kinect, which contains color pictures with a resolution of $640 \times 480$ and corresponding depth maps. The dataset contains 120K training samples and 654 testing samples~\cite{2014Depth}. We utilize a 50K subset for training with the method of Levin \textit{et al}.~\cite{levin2004colorization} to fill the missing depth values like Alhashim \textit{et al}.~\cite{alhashim2018high}. In the preprocessing stage, the size of the input image is scaled to half of the original size, which is $320 \times 240$, and then a center cropping with a resolution of $304 \times 228$ is applied. Considering the size of the output through the network being half of the input size, we downsample the depth map of the training set to $114 \times 152$, while the size of the depth map in the testing set is still $304 \times 228$. During the training process, the reciprocal of the depth is adopted to suppress the situation where the loss increases with the depth value increasing. When testing, we need to take the reciprocal of the network output to get the true depth value, and then resize it to the same size as the input image.

The KITTI dataset contains outdoor scenes with images of resolution about $375 \times 1241$ captured by cameras and depth sensors in a driving car~\cite{geiger2013vision}. We train our model to use a subset of approximately 26K left viewed images with 697 testing pictures divided by Eigen \textit{et al}.~\cite{2014Depth}. We directly exploit the sparse data collected by LIDAR sensor with a depth limit of 80 meters and mask the pixels without depth values when calculating the loss. We set the random crop size of the training set image to $385 \times 513$ and downsample the corresponding depth map by half to $193 \times 257$ while applying the full image when testing, and resize the network output to the full image size. Note that the real depth value is applied during training on KITTI dataset instead of the inverse depth, which is different from the NYU-Depth V2 dataset.

\subsection{State-Of-The-Art Comparisions}
\textit{1)NYU-Depth V2: } Table.\ref{Tab1} shows the performance comparison among our proposed method and other related methods on NYU-Depth V2 dataset. It can be seen from Table.\ref{Tab1} that the proposed method is competitive with other methods. Among the six recognized metrics, our method is the highest in the four of them, except for $\delta \textless 1.25^3$, we are only 0.002 slightly lower than Zhang \textit{et al}.~\cite{zhang2019pattern}, which proves that our method has outstanding performance on NYU-Depth V2 dataset. Our method is slightly behind in RMSE.
We analyze the difference among our approach, Zhang \textit{et al}.~\cite{zhang2019pattern}, Wang \textit{et al}.~\cite{wang2020sdc}, Su \textit{et al}.~\cite{su2020monocular} and Wang \textit{et al}.~\cite{wang2020cliffnet}. We find that the other four methods additionally added semantic information. Specifically, Zhang \textit{et al}.~\cite{zhang2019pattern} and Wang \textit{et al}.~\cite{wang2020sdc} utilized a multi-task training approach so that the depth estimation task shared encoders with the semantic segmentation task, and the encoding process appended semantic information constraints for the depth estimation task. Su \textit{et al}.~\cite{su2020monocular} applied a channel attention to enhance the semantic information among channels. Wang \textit{et al}.~\cite{wang2020cliffnet} employed a scene classification model to train the depth estimation model, which was also a form of adding semantic information. Semantic information is beneficial for depth estimation tasks, especially for edge pixels of objects. Although their methods perform slightly better than our method in RMSE, a metric that focuses on pixel accuracy, our method integrates detailed information, spatial information, and samples' information, and thus outperforms these methods for the rest of the metrics.

\begin{figure*} 
\centering    
\includegraphics[width=170mm]{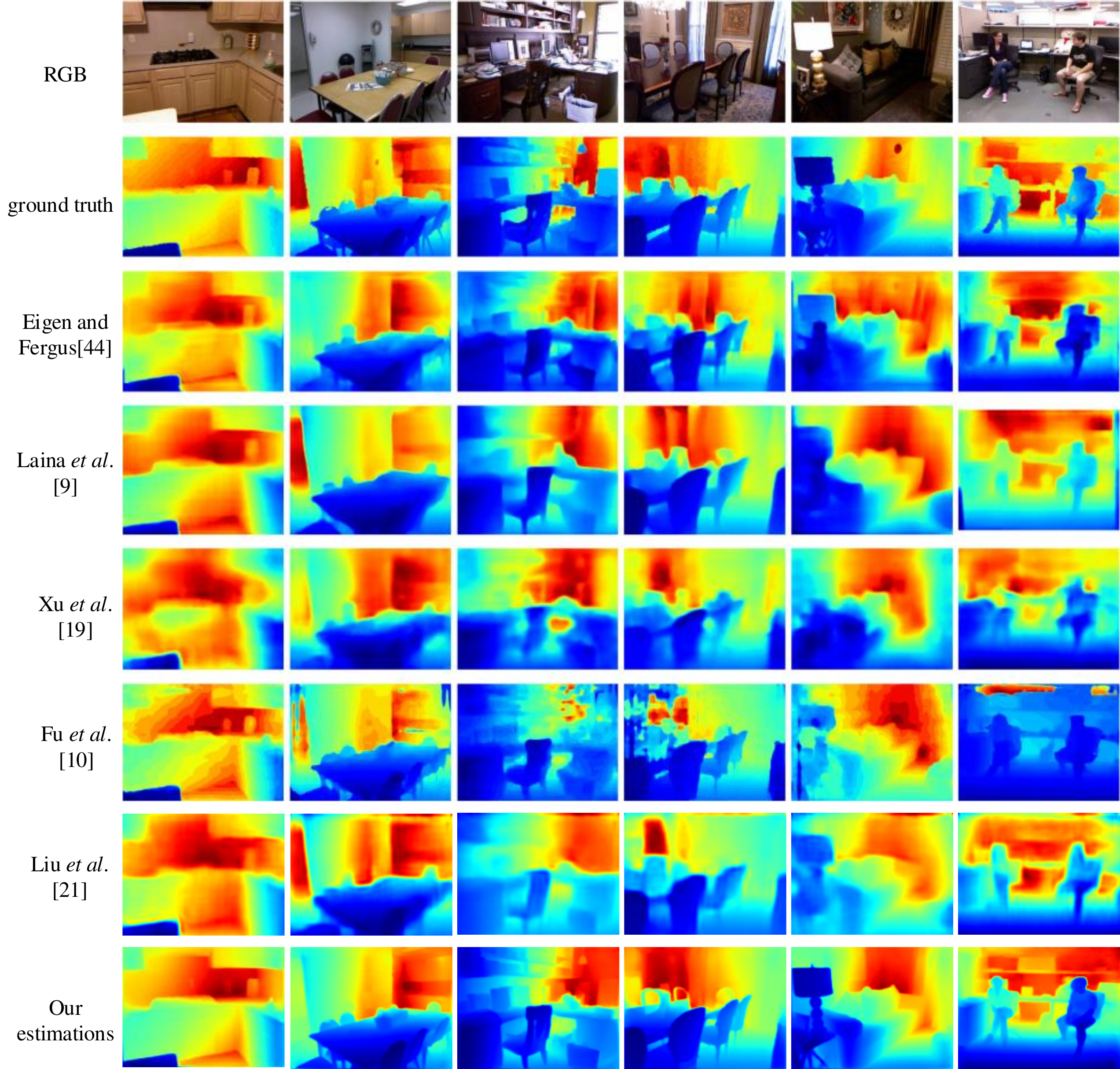}\\    
\caption{Comparison of predicted results on NYU-Depth V2 dataset. From top to bottom: RGB input, ground truth, Eigen and Fergus.~\cite{eigen2015predicting}, Laina \textit{et al}.~\cite{laina2016deeper}, Xu \textit{et al}.~\cite{xu2017multi}, Fu \textit{et al}.~\cite{fu2018deep}, Liu \textit{et al}.~\cite{liu2019fully} and the results of our method.}
\label{Fig6}    
\end{figure*}

The qualitative comparison on NYU-Depth V2 dataset among our method and other methods is shown in Fig.\ref{Fig6}. It demonstrates that our method is superior to other methods in recovering details and planar regions. For example, our method predicts
accurate geometric details for the gap of the table in the lower left corner of the second column, the computer on the table in the third column, and the lamp in the fifth column. For the planar regions (the corner of the wall in the first column, the desk in the fourth column, and the desk on the left side of the sixth column), our method also generates better results.

\begin{figure*}  
\centering    
\includegraphics[width=170mm, ]{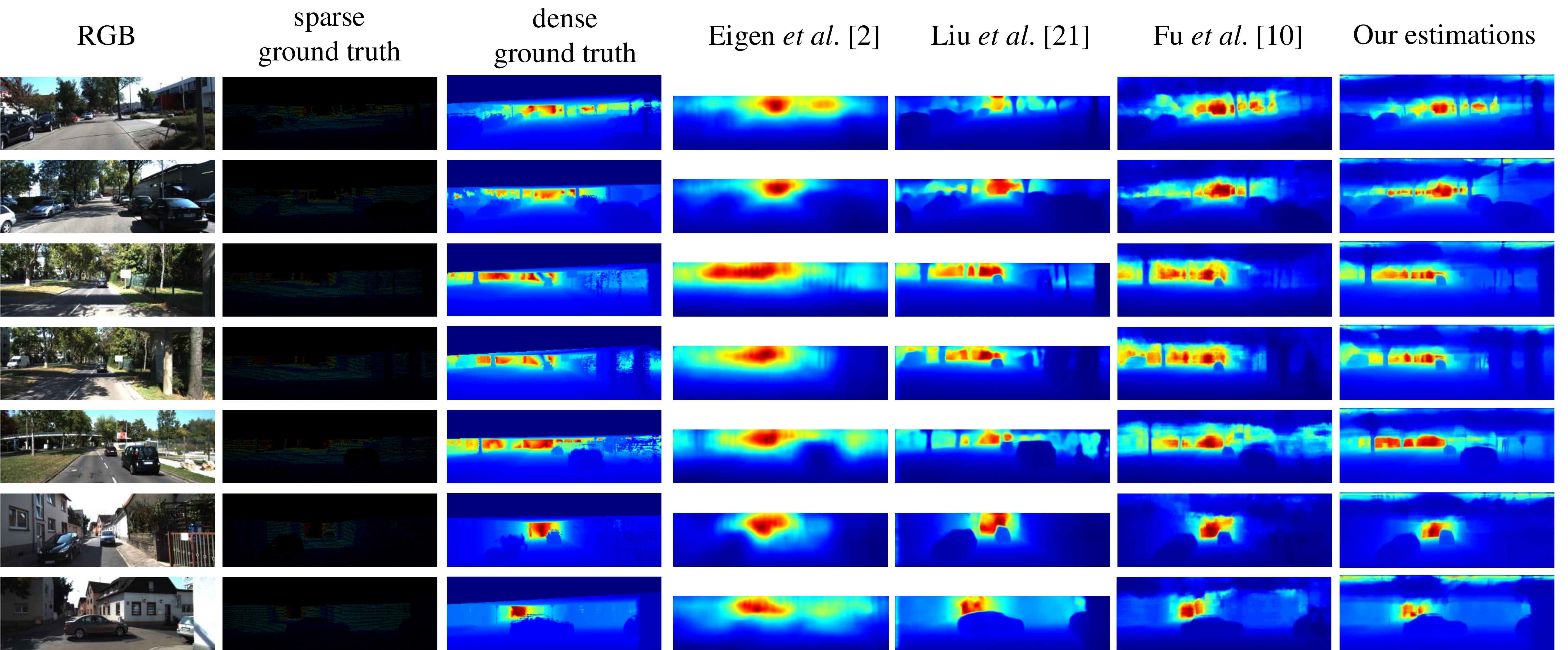}\\    
\caption{Comparison of predicted results on KITTI dataset. From left to right: RGB input, ground truth, Eigen \textit{et al}.~\cite{2014Depth}, Liu \textit{et al}.~\cite{liu2019fully}, Fu \textit{et al}.~\cite{fu2018deep}, and the results of our network.}
\label{Fig7}   
\end{figure*} 

\textit{2)KITTI: } Table.\ref{Tab2} shows the results of our proposed method on the challenging KITTI dataset. From the data in the table, it is obvious that our method has achieved the best results in the all of five metrics compared with other methods. Since our method is directly supervised on KITTI dataset, fairly, we merely compare to Fang \textit{et al}.~\cite{fang2020towards} with the best supervised metrics listed in the paper. The qualitative comparison on KITTI dataset among our method and other methods is shown in Fig.\ref{Fig7}. It is worth noting that Eigen \textit{et al}.~\cite{2014Depth} and Liu \textit{et al}.~\cite{liu2019fully} used the effective part of ground truth data, which was the lower 2/3 part of the image. Meanwhile, Liu \textit{et al}.~\cite{liu2019fully} and Fu \textit{et al}.~\cite{fu2018deep} applied the dense depth maps to train the network. While the input of our network is the full image, and the original sparse data is utilized as the ground truth. There are two advantages: 1) The steps of image preprocessing can be reduced, and the original image obtained by the camera can be directly applied. 2) The depth ground truth gotten by LIDAR can be straightly exploited without processing into a dense map. Compared with existing methods, our method can provide a depth map with clearer outlines and more detailed information. For example, in Fig.\ref{Fig7}, from the first row to the seventh row, compared to other methods, our method can clearly predict the outline of the trees.

\subsection{Ablation Studies}
\label{Ablation Studies}
 To prove the effectiveness of the three innovations in this paper, we introduce the result in detail via ablation experiments in this section.
 Like Su \textit{et al}.~\cite{su2020monocular}, we also avail of a NYU-Depth V2 subset for experiments. The subset contains 1449 finely labeled color images and their corresponding depth maps, including 795 images as the training dataset and 654 images as the testing dataset. In the ablation experiments, except for training 40 epochs, the hyperparameters and data augment methods are the same as the training and testing on the large dataset introduced earlier. The results of the ablation experiments are shown in Table.\ref{Tab3}. 
 
 From Table.\ref{Tab3}, it is obvious that the performance is gradually improved by incorporating the USF module, EDA module, and batch-loss. More specifically, after adding the USF module, the metrics are improved compared to the baseline, where REL is reduced by 1.3\% and $\delta \textless 1.25$ is increased by 1.1\%, which shows that the fusion of multi-stage features is effective for the results of depth estimation. It is worth noting that RMSE is increased by 1.1\% and is not reduced as expected. The reason may be that the feature fusion improves the overall accuracy while the accuracy for each pixel is still poor, especially the prediction mistakes of far distance can directly produce large errors. Thus, we consider adding the spatial attention mechanism into the network for mining the significant spatial information to alleviate the above problems.
 
\begin{figure} 
\centering    
\includegraphics[width=87mm]{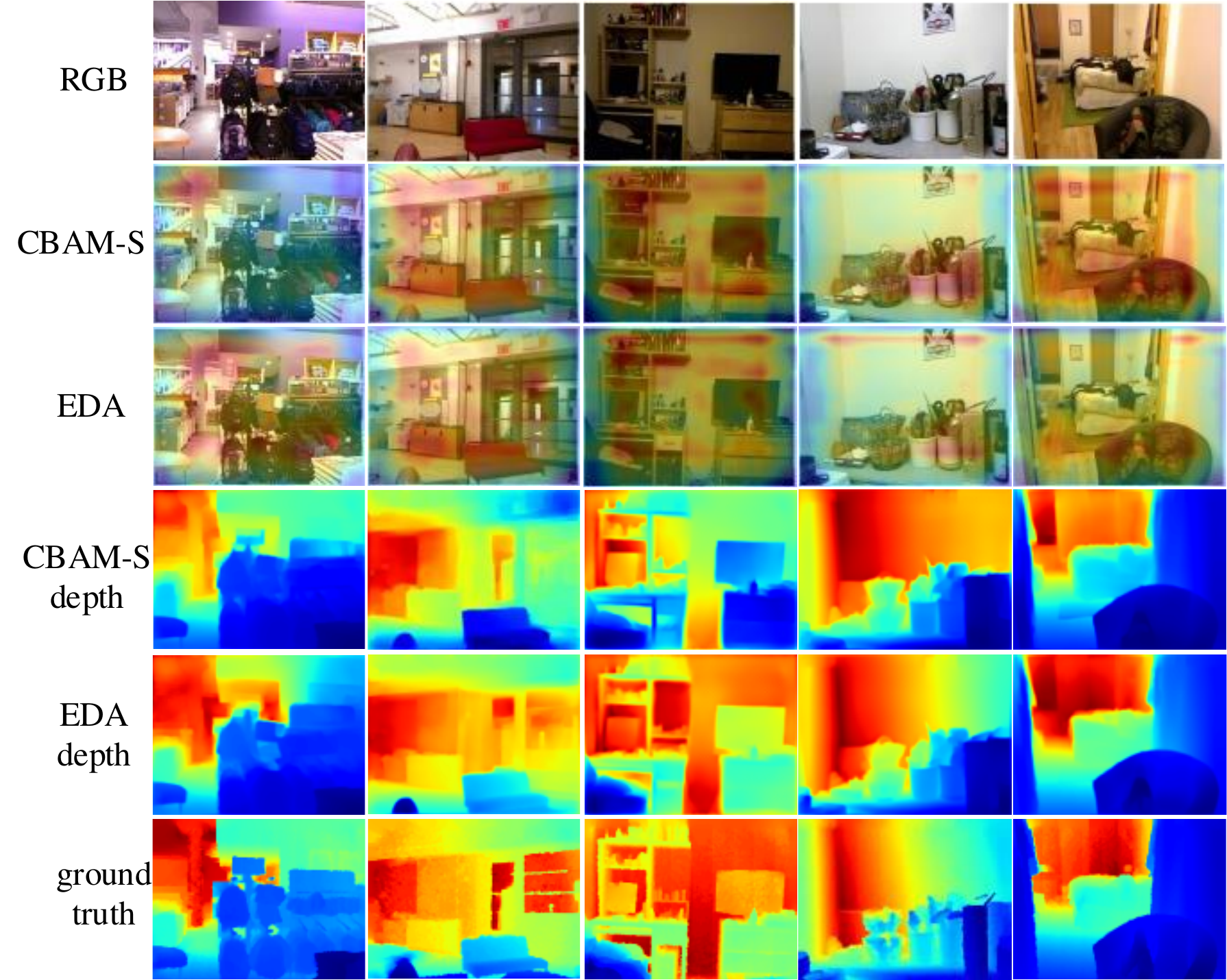}\\    
\caption{Comparison of EDA module and CBAM-S. From top to bottom: RGB input, CBAM-S visualization, EDA module visualization, the depth map of CBAM-S, the depth map of EDA module, and ground truth. The depth maps obtained by CBAM-S and EDA module are the predicted results of the baseline+ (CBAM-S/EDA module).}
\label{Fig4}    
\end{figure}

 As mentioned in Section~\ref{B}, CBAM-S is employed to carry out experiments. The result is shown in the fourth row in Table.\ref{Tab3}. After adding CBAM-S, the performance has been further improved, where REL is reduced by 0.8\% and $\delta \textless 1.25$ is increased by 0.9\%, while RMSE is still increased. It can be illustrated that spatial attention is meaningful for supplementing detailed information, while the information (especially far distance) obtained by CBAM-S is not enough and the noise in the feature map also causes an increase in RMSE. For the above two reasons, neither maximum pooling nor average pooling in CBAM-S can weaken the negative effects. When utilizing our EDA module to replace CBAM-S, REL is reduced by 1.3\%, and $\delta \textless 1.25$ is increased by 2.4\%. More obviously, RMSE is reduced by 4.9\%. Meanwhile, compared with USF alone, REL is reduced by 2.1\%, RMSE is reduced by 2.5\%, and $\delta \textless 1.25$ is increased by 3.3\%, which is an apparent improvement. It can be seen that the subtraction operation in the EDA module can lower the impact of noise. With the high response regions being crippled, the network is forced to mine more details, increasing the probability of distant objects being paid attention to and enriching the diversity of features, to achieve a considerable effect. 
 In addition to the comparison of metrics, we also make a qualitative comparison between CBAM-S and EDA module, which is shown in Fig.\ref{Fig4}. To demonstrate a clearer visual result, we apply the large dataset of NYU-Depth V2 to the experiment. In Fig.\ref{Fig4}, compared with CBAM-S, our EDA module has a better predictive ability for further objects. In particular, the depth map of the EDA module in the red area is closer to the ground truth, causing the error to become smaller and RMSE being reduced substantially are reasonable.
 
 In the above experiments, the loss functions of the network are $L_{grad}$, $L_{SSIM}$, and $L_1$loss. For each module added, an auxiliary loss in the form of $L_1$ needs to be added to calculate the deviation between the feature map and the ground truth. At last, based on complete network structure, batch-loss is utilized to replace $L_1$ loss, with the auxiliary loss remaining $L_1$ loss, and RMSE is straight reduced by 4.5\%. It is apparent that strengthening the learning of hard samples is meaningful to further improve the generalization ability of the model. From Tabel.\ref{Tab3}, when all innovation points are appended to the experiment, compared with baseline, REL is reduced by 3.7\%, RMSE is reduced by 5.9\%, and $\delta \textless 1.25$ is increased by 5.5\%, the optimal results being achieved. For the monocular depth estimation, the feature fusion strategy, the spatial attention mechanism, and utilizing the contribution of the hard examples are all effective.

\section{Conclusion}
 In this paper, we propose a Multi-scale Features Network for monocular depth estimation. The network only exploits RGB images to train models and predict results with the end-to-end method. The proposed network consists of two effective modules and a novel loss function. These two modules named the EDA module and the USF module are used to refine features and provide more detailed information for the decoder part. Specifically, The EDA module considers the influence of noise and mines sufficient spatial information, and the USF module is utilized to fuse information from multiple stages. The novel batch-loss function is employed to strengthen the contribution of the hard examples, which is beneficial to improve the performance of the model. The validations on NYU-Depth V2 dataset and KITTI dataset prove the capability of our proposed method. In future work, how to achieve a better balance between calculation time and prediction accuracy will be conducted.

\section{acknowledgment}
This work is supported by Major Science and technology innovation engineering projects of Shandong Province (2019JZZY010128), National Natural Science Foundation of China (No. 61973066), Distinguished Creative Talent Program of Liaoning Colleges and Universities (LR2019027).
\bibliographystyle{IEEEtran}

\end{document}